\pdfoutput=1
\documentclass[sigconf]{acmart}

\AtBeginDocument{%
  }

\copyrightyear{2023}
\acmYear{2023}
\setcopyright{acmlicensed}\acmConference[MM '23]{Proceedings of the 31st ACM International Conference on Multimedia}{October 29-November 3, 2023}{Ottawa, ON, Canada}
\acmBooktitle{Proceedings of the 31st ACM International Conference on Multimedia (MM '23), October 29-November 3, 2023, Ottawa, ON, Canada}
\acmPrice{15.00}
\acmDOI{10.1145/3581783.3612101}
\acmISBN{979-8-4007-0108-5/23/10}

\usepackage{multirow}
\usepackage{enumitem}

\begin{document}

%%%%%%=====Defined by Yi====%%%%%%
\newcommand{\myet}{\emph{et~al.}}
\newcommand{\myeg}{\emph{e.g.}}
\newcommand{\myie}{\emph{i.e.}}
\newcommand{\myec}{\emph{etc}}
\newcommand{\by}[1]{\textcolor{magenta}{#1}}
%%%%%%=====Defined by Yi====%%%%%%

\title{Your Negative May not Be True Negative:\\Boosting Image-Text Matching with False Negative Elimination}

\author{Haoxuan Li}
\affiliation{%
  \institution{University of Electronic Science and Technology of China}
  \city{Chengdu}
  \country{China}
}
\email{lhx980610@gmail.com}

\author{Yi Bin}
\authornote{Yi Bin is the corresponding author (Email: yi.bin@hotmail.com).}
\affiliation{%
  \institution{National University of Singapore}
  \country{Singapore}
}
\email{yi.bin@hotmail.com}

\author{Junrong Liao}
\affiliation{%
  \institution{University of Electronic Science and Technology of China}
  \city{Chengdu}
  \country{China}
}
\email{charliel114514@gmail.com}

\author{Yang Yang}
\affiliation{%
  \institution{University of Electronic Science and Technology of China}
  \city{Chengdu}
  \country{China}
}
\email{yang.yang@uestc.edu.cn}

\author{Heng Tao Shen}
\affiliation{%
  \institution{University of Electronic Science and Technology of China}
  \city{Chengdu}
  \country{China}
}
\email{shenhengtao@hotmail.com}

\renewcommand{\shortauthors}{Li et al.}

\begin{abstract}

  Most existing image-text matching methods adopt triplet loss as the optimization objective, and choosing a proper negative sample for the triplet of <anchor, positive, negative> is important for effectively training the model, \myeg{}, hard negatives make the model learn efficiently and effectively. However, we observe that existing methods mainly employ the most similar samples as hard negatives, which may not be true negatives. In other words, the samples with high similarity but not paired with the anchor may reserve positive semantic associations, and we call them \textit{false negatives}. Repelling these false negatives in triplet loss would mislead the semantic representation learning and result in inferior retrieval performance. In this paper, we propose a novel \textbf{False Negative Elimination (FNE)} strategy to select negatives via sampling, which could alleviate the problem introduced by false negatives. 
  Specifically, we first construct the distributions of positive and negative samples separately via their similarities with the anchor, based on the features extracted from image and text encoders. Then we calculate the false negative probability of a given sample based on its similarity with the anchor and the above distributions via the Bayes' rule, which is employed as the sampling weight during negative sampling process. Since there may not exist any false negative in a small batch size, we design a memory module with momentum to retain a large negative buffer and implement our negative sampling strategy spanning over the buffer.
  In addition, to make the model focus on hard negatives, we reassign the sampling weights for the simple negatives with a cut-down strategy. 
  The extensive experiments are conducted on Flickr30K and MS-COCO, and the results demonstrate the superiority of our proposed false negative elimination strategy. The code is available at \by{\url{https://github.com/LuminosityX/FNE}}.

\end{abstract}



%\ccsdesc[100]{Computing methodologies~Computer vision}


\maketitle

\begin{figure}[ht]
  \centering
  \includegraphics[width=\linewidth]{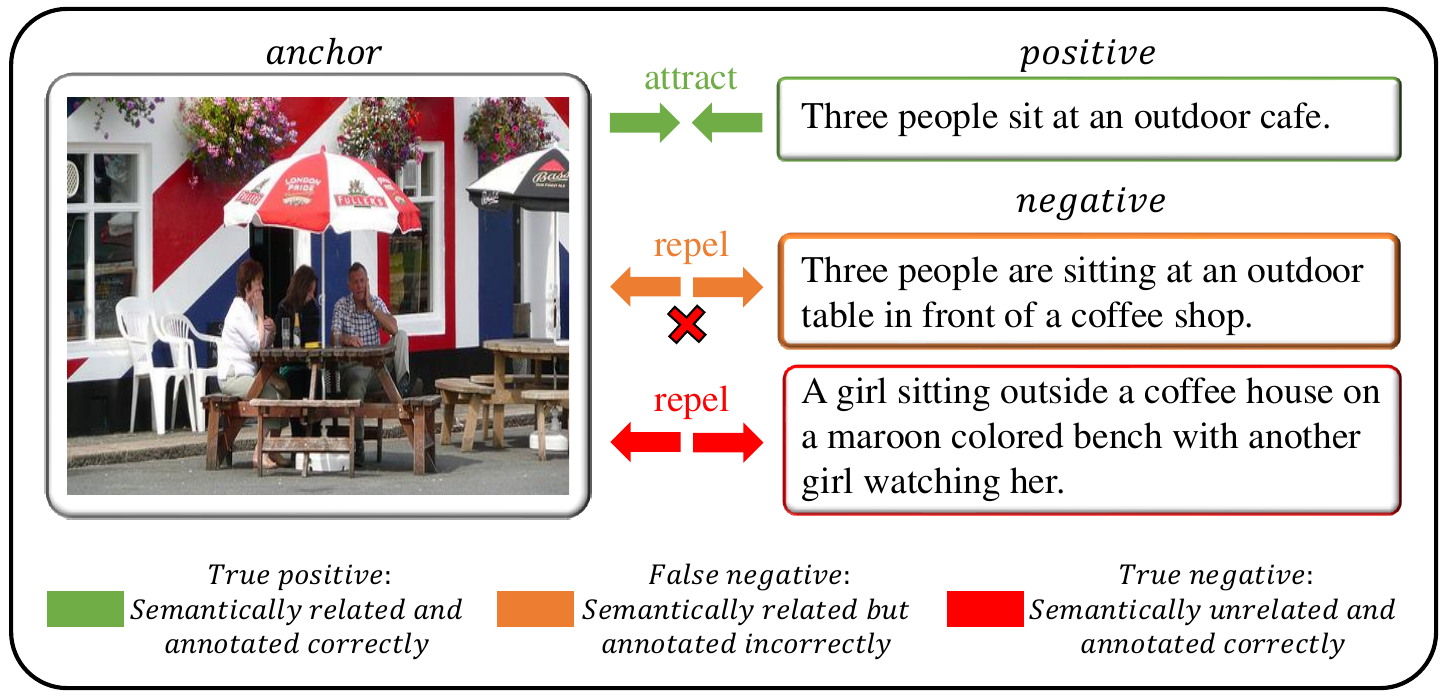}   %% img_1_3
  \caption{Illustration of a triplet. The triplet consists of an anchor, a positive sample, and a negative sample, and tries to attract the positive samples and repel the negative samples. However, the negatives may not be true negatives, such as the negative sample in the orange box sharing the same semantics with the image, which should not be considered as negative and is called \textit{false negative} in this paper.}
\label{fig:fig1}
\end{figure}

\section{Introduction}
With the growing power of deep learning methods and the accessibility of various types of data, the field of multi-modal analysis has gained widespread attention. Among them, jointly understanding and exploring of vision and language are the most crucial problem. In this paper, we focus on the task of image-text matching, which aims to search the most relevant image (or text) with a query text (or image). This task is a fundamental research problem in the field of vision-language understanding and is beneficial to numerous tasks such as image captioning~\cite{feng2019unsupervised,bin2021multi}, visual question answering (VQA)~\cite{anderson2018bottom,peng2021progressive}, and others~\cite{bin2022non, xu2023multi,deng2018visual,bin2017adaptively}. Despite the exciting progress that has been made in this area, jointly learning the image and text representations for more accurate semantic alignment remains challenging due to the inherent heterogeneous gap between visual and textual modalities.

Existing works mainly adopt two kinds of methods to learn the correspondence between visual and textual representations: the Visual Semantic Embedding (VSE) methods~\cite{frome2013devise, faghri2017vse++, chen2021learning, li2023selectively} and the Cross-Attention (CA) methods~\cite{lee2018stacked, wei2020multi, qu2021dynamic, zhang2022negative}. Specifically, the VSE methods separately embed the whole image and text into a common semantic space by two independent networks. As prior works, the methods~\cite{frome2013devise, kiros2014unifying} employ Convolutional Neural Networks (CNNs) and Recurrent Neural Networks (RNNs) to extract visual and textual features for representation learning, respectively. More recently, some works~\cite{qu2020context, chen2021learning, li2023selectively} extract words and regions features by BERT~\cite{devlin2018bert} and Faster R-CNN~\cite{ren2015faster} to explore the fine-grained associations between images and sentences. However, the VSE methods is limited in terms of matching performance due to that it only explores the semantic interactions within each modality, resulting in lacking of sufficient modal interaction. Therefore, the CA methods are proposed to learn rich inter-modal interactions utilizing a cross-attention mechanism. For instance, SCAN~\cite{lee2018stacked} explores all latent alignments by attending words to each regions or attending regions to each word. While the cross-attention operation has brought significant performance improvements, and also induced huge computational costs.

As we know, most existing image-text matching methods implement triplet loss as the optimization objective to train the model. 
For image-text matching, a triplet typically consists of an anchor query image\footnote{To make the paper clear, here we describe the image-to-text matching process as example, and vice versa for text-to-image matching.}, a positive text, and a negative text, which formulates a positive pair and a negative pair.
As shown in Figure~\ref{fig:fig1}, the optimization objective of triplet loss is to narrow the distance between anchor and positive samples and push away the negative samples in the common space. Through such attraction and repulsion for different kinds of samples, the model could learn the implicit semantic associations and result in precise image-text matching. Naive triplet loss in image-text matching~\cite{frome2013devise} only leverages the negative samples in current mini-batch, and cannot exhaustively explore the ones in the whole dataset. 
With the introduction of the hard negative mining strategy~\cite{faghri2017vse++}, which only considers the negative sample with the highest matching score for optimization to avoid redundant samples easy to distinguish, and makes the model learn more discriminative representations. Then many works attempt to devise and employ various hard negative mining strategies and significantly improves the image-text matching performance~\cite{chen2020adaptive, qu2021dynamic, zhang2022negative, lee2018stacked}. However, some works point out that the hardest negatives may make the distance metrics fail to capture the semantics and lead to bad local minima\cite{schroff2015facenet, xuan2020improved, xuan2020hard, yu2018correcting, oh2016deep}, and propose semi-hard triplet mining and easy positive mining to alleviate this issue. We also observe that some negative samples actually share the same semantics with the anchor and positive samples. Besides, due to the semantic diversity and flexibility, there could exist multiple texts to describe one identify image, and vice versa. While hard negative mining strategy attempts to push them all away, excepting the ones annotated positive with the anchor in the dataset, by the triplet loss.
Obviously, repelling such negatives with positive semantics, called \textit{\textbf{false negative}} in this paper, may mislead the model learning and result in inferior semantic representations. 

To alleviate the problems caused by false negatives, we propose a novel False Negative Elimination strategy, namely FNE, to sample the negatives with different weights. Intuitively, a negative sample holds higher similarity with an anchor, it is more likely to share the similar or same semantics with the anchor, which means higher confidence to be a \textit{false negative}. With such assumption, we first construct two distributions based on the similarities of positive pairs (all the anchors and corresponding positive samples) and negative pairs. Based on the similarity distributions, we can easily calculate the probability that a given example is a false negative by Bayes' rule. We then implement a weighted-sampling strategy by assigning lower weights to avoid the false negative to be included in the triplet loss. However, limited by the GPU memory, mini-batching technique is commonly-used during training, but there may not exist any false negatives with a small batch-size. To make more samples accessible in the sampling process, we introduce a novel momentum memory module to retain a large buffer of negative samples, and improve the occurrence frequency of false negatives.
Previous works~\cite{xuan2020hard,zhang2022negative,chen2020adaptive} point out that the easy negatives (obviously reverse farther distance with the anchor than positives) are redundant and cannot provide much information in the triplet loss optimization. We therefore introduce a simple cut-down strategy, decreasing the sampling probability of such easy negatives, to make the model focus on hard negatives and learn better semantic representations. Extensive experiments have been conducted on MS-COCO and Flickr30K, and verified the effectiveness of our proposed approach.
The main contributions are as follows:
\begin{itemize}[leftmargin=*]
    \item We propose a novel \textbf{False Negative Elimination (FNE)} strategy to effectively alleviate the false negative problem during image-text matching training. Our FNE implements weighted-sampling for hard negatives with posterior probability, and attempts eliminate false negatives by assigning lower sampling weights.
    \item To make more samples available during sampling process, we introduce a momentum memory module to enlarge the sampling pool from a small mini-batch to a large buffer. This buffer retains more negatives and results in more occurrence of false negatives.
    \item Extensive experiments have been conducted on two commonly-used datasets, MS-COCO and Flickr30K in specific. The experimental results demonstrate that our proposed methods outperforms all the SOTAs and verify its effectiveness.
\end{itemize}

%%%%%%%%%%%%%%%%%==Figure Block==%%%%%%%%%%%%%%%
\begin{figure*}[h]
\begin{center}
\includegraphics[width=0.98\textwidth]{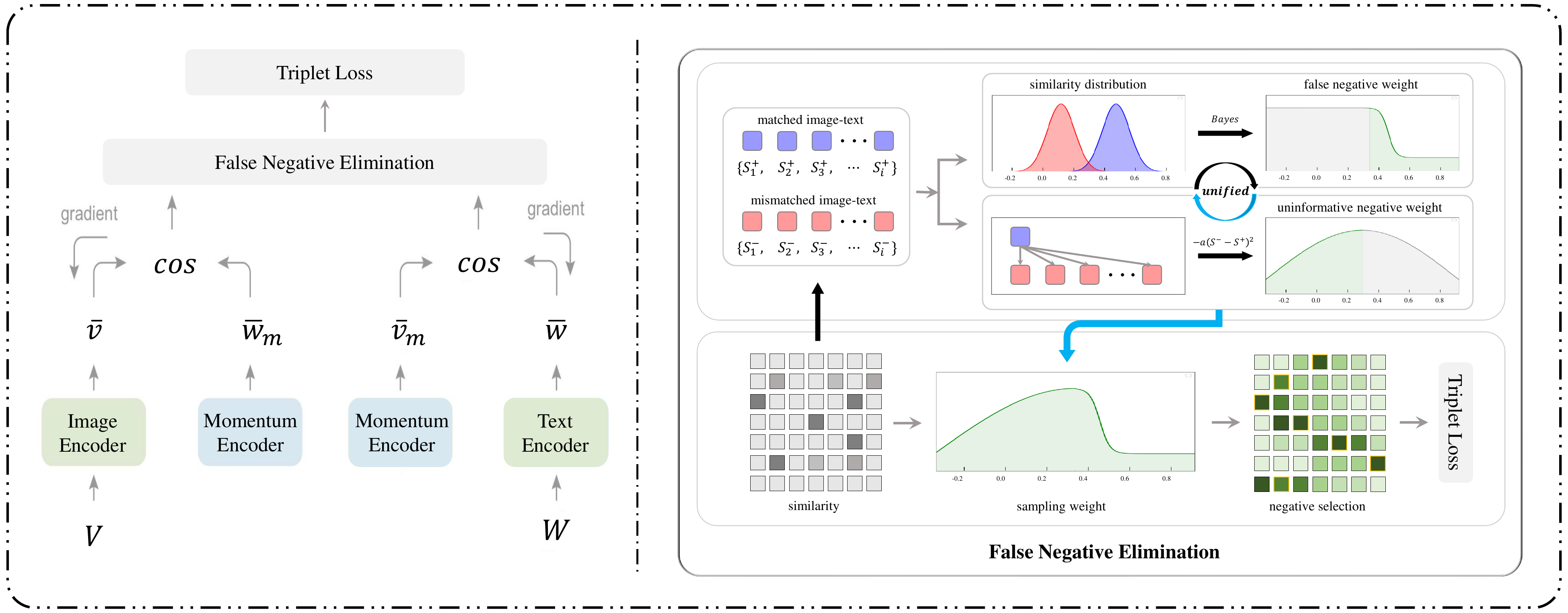}
%7.16
\end{center}
\caption{Illustration of the pipeline of our proposed False Negative Elimination (FNE) strategy and model framework. The model framework consists of the original feature encoder and its corresponding Momentum Memory Module. The proposed FNE strategy assigns sampling weights to negative samples based on the probability of being false negatives, whereby higher probabilities result in lower weights. Finally, this leads to a more accurate and discriminative feature representation.
}
% \vspace{-10pt}
\label{fig:overview}
\end{figure*}
%%%%%%%%%%%%%%%%%==Figure Block==%%%%%%%%%%%%%%%

\section{Related Works}
\subsection{Image-Text Matching}
\label{sec:rel_1}

As pointed in~\cite{li2023selectively, li2022multi}, existing image-text matching methods can be divided into two categories, Visual Semantic Embedding method~\cite{frome2013devise, faghri2017vse++, chen2021learning, li2023selectively, wang2023multilateral}, and Cross-Attention method~\cite{lee2018stacked, wei2020multi, qu2021dynamic, zhang2022negative, ge2023cross}.

\noindent \textbf{Visual Semantic Embedding:} The VSE methods independently projecting the entire image and text into a common embedding space with a two-branch neural network. Therefore, these methods can increase the inference speed by cache embedding, making them efficient in real-world scenarios. The first visual semantic embedding model was proposed by Frome \textit{et al.}~\cite{frome2013devise}, which employed the CNN and Skip-Gram~\cite{mikolov2013efficient} for visual and language feature extraction. Faghri \textit{et al.}~\cite{faghri2017vse++} proposed the VSE++ method, which integrated online hard negative mining strategy into the triplet loss function. Chen \textit{et al.}~\cite{chen2021learning} proposed a Generalized Pooling Operator, which generates the best pooling strategy by learning different weights for each ranking dimension. Recently, Zheng \textit{et al.}~\cite{li2023selectively} proposed the well-scaled RVSE++ method, which solves the gradient vanishing problem and achieves the state-of-the-art performance.

\noindent \textbf{Cross-Attention:} The CA methods explore the fine-grained semantic correspondences between images and texts through the cross-attention mechanism. A stacked cross attention (SCAN) model was proposed by Lee \textit{et al.}~\cite{lee2018stacked}, which measures the image-text similarity by selectively cross aggregating regions and words. Later, Wei \textit{et al.}~\cite{wei2020multi} introduced the self-attention and cross-attention mechanisms to model both intra-modal and inter-modal interactions. Qu \textit{et al.}~\cite{qu2021dynamic} introduced a dynamic routing mechanism for model embedding, which dynamically selects the embedding path based on the input and designed four types of interaction cells for intra-modal and inter-modal interaction. Recently, Zhang \textit{et al.}~\cite{zhang2022negative} proposed the NAAF method based on SCAN method, which further refines the calculation of similarity by considering the negative impact caused by mismatched fragments. However, during inference, the cross-attention method requires calculating cross-attention on all visual and textual data, resulting in significantly lower efficiency compared to the VSE methods. This makes it unsuitable for large-scale vision-language retrieval.

\subsection{Negative Samples Selection}

The selection of negative samples is critical and has been widely studied in various fields, including retrieval and recommendation~\cite{ding2023personalized,ding2021leveraging,wang2023multilateral} In contrastive learning, PIRL~\cite{misra2020self} first expanded the selection range of negative samples and increased the probability of selecting hard negative samples by maintaining a memory bank. Then, MoCo~\cite{he2020momentum} introduced a momentum model based on the memory bank, further enhancing its scalability. For image-text matching, early works~\cite{frome2013devise} utilized all negative samples with triplet loss for training. Faghri \textit{et al.}~\cite{faghri2017vse++} proposed the triplet loss with online hard negative mining strategy. Most image-text matching methods adopt this selection strategy to optimize their models, and achieves remarkable improvements. Later, the AOQ method~\cite{chen2020adaptive} employed a pre-trained model to search for the hardest sample pairs within the entire dataset before training and Wei \textit{et al.}~\cite{wei2021universal} proposed a universal weighting framework to assign larger weight to harder sample, which further improved retrieval performance. However, these works only focus on selecting the hardest negative samples to learn more discriminative feature representations, while ignoring the semantic diversity in both images and text. The semantic diversity results in the existence of sample pairs within negative samples that are actually matching, which are referred to false negatives.

\section{The Proposed method}
The overall pipeline of our proposed False Negative Elimination (FNE) strategy and model framework are illustrated in Figure~\ref{fig:overview}. The model framework consists of two feature encoder and its corresponding momentum memory module. When calculating the triplet loss function, we apply our proposed FNE strategy to select negative samples to reduce the presence of false negatives. In this section, we will elaborate on them in detail. Firstly, we introduce the way to extract visual and textual representations in Sec~\ref{sec:sec3_1}. Then, we describe the details of our proposed False Negative Elimination strategy in Sec~\ref{sec:sec3_2}. Finally, we introduce the momentum memory module and the final objective function in Sec~\ref{sec:sec3_3} and Sec~\ref{sec:sec3_4}.

\subsection{Feature Representations}
\label{sec:sec3_1}
%To obtain a more compatible embedding space
\noindent \textbf{Image Representation.} We adopt the ViT model~\cite{dosovitskiy2020image} based on transformer~\cite{vaswani2017attention} structure to extract the image features. The ViT model divides a given image into patches and employs self-attention mechanism to better learn the spatial context information of the image, ultimately obtaining the patch features of the image. To better align the semantics between visual and text modalities, following previous works~\cite{qu2020context, qu2021dynamic}, we project the patch features into a common semantic space through a fully-connection layer. Finally, the learned patch features could be denoted as $\textbf{V}=\{\textbf{v}_i|i=1, ..., m, \textbf{v}_i\in\mathbb{R}^d\}$, where $m$ indicates the number of patches for each image.

\noindent \textbf{Text Representation.} To process the input texts, we follow recent trends in the Natural Language Processing community and utilize a pre-trained BERT~\cite{devlin2018bert} model, BERT-based in specific, to extract contextual word representations. 
Similarly, we employ fully-connection layers to project the extracted word features into a common semantic space, and obtain $\textbf{W}=\{\textbf{w}_j|j=1, ..., l, \textbf{w}_j\in\mathbb{R}^d\}$, where $l$ denotes the number of words in the sentence.

\begin{equation}
    \bar{\mathbf{v}} = \frac{1}{m} \sum_{i=1}^m \mathbf{v}_i, \quad  \bar{\mathbf{w}} = \frac{1}{l} \sum_{j=1}^l \mathbf{w}_j.
\end{equation}
To facilitate subsequent optimization of the loss function and the calculation of final similarity, we adopt an average pooling operation to obtain global features for both images and texts.

\subsection{False Negative Elimination Strategy}
\label{sec:sec3_2}
Once the global features of both images and texts are obtained, the triplet loss will be employed to supervise the alignments between visual and textual semantics. A metric function is first applied to measure the distance or similarity between the paired image $\bar{\mathbf{v}}$ and sentence $\bar{\mathbf{w}}$, both for positive and negative pairs. Here we implement the commonly-used cosine similarity as:
\begin{equation}
\label{equ:similarty}
   s(\mathbf{\bar{v}}, \mathbf{\bar{w}})=\frac{\mathbf{\bar{v}}^T\mathbf{\bar{w}}}{||\mathbf{\bar{v}}||||\mathbf{\bar{w}}||}.
\end{equation}
Subsequently, the triplet loss function is employed to achieve the attraction of positive pairs and the repulsion of negative pairs, which could be formulated as:
\begin{equation}
\label{equ:triplet}
\begin{aligned}
        \mathbf{\mathcal{L}_{tri}} = &[margin - s(\mathbf{\bar{v}}, \mathbf{\bar{w}}) + s(\mathbf{\bar{v}}, \mathbf{\bar{w}^-})]_+ \\ + \ &[margin - s(\mathbf{\bar{v}}, \mathbf{\bar{w}}) + s(\mathbf{\bar{v}^-}, \mathbf{\bar{w}})]_+ ,
\end{aligned}
\end{equation}
where $margin$ is a constraint hyper-parameter, implying that a negative should stay further away from the anchor than the positive with $margin$, or the triplet will be penalized. $[\cdot]_+=max(0,\cdot)$. $\mathbf{\bar{w}^-}$ denotes the hard negative text embedding and $\mathbf{\bar{v}^-}$ denotes the hard negative image embedding in a mini-batch.

Most existing hard negative mining strategies choose the most similar examples, \myie{}, highest similarity with anchor, to construct the <anchor, positive, negative> triplet. As aforementioned, such hard negatives with very high similarities may share the same semantics and actually match with the anchor, we call them \textit{\textbf{false negative}} in this paper, due to the semantic diversity. The triplet loss, however, still attempts to push them away from the anchor, which may make the model confused to learn the semantic representations and then lead to inferior image-text matching performance. The ideal way to address this problem would be directly removing these false negative samples during triplet loss calculation. But it is hard to automatically and totally exclude them from the hard negatives, without explicit label to indicate whether a sample is a false negative or true negative. Therefore, we propose a novel negative sampling strategy, termed false negative elimination (FNE), which tries to alleviate this problem via deceasing the occurrence probability of false negatives. In specific, given a negative sample, the propose FNE strategy attempts to estimate the posterior probability that it to be a false negative, based on the Bayes' rule and prior probability distributions of positive and negative samples.

As aforementioned, false negatives refer to the samples that are defined as negative samples in the dataset, but actually match with the anchor in semantics. 
Since we try to employ the posterior probability to estimate \textit{how likely a negative sample that to be a false negative}, which is equivalent to estimate how likely it matches with the anchor, \myie{}, to be a positive. Towards this end, we need to investigate the statistical property of positive pairs based on the match scores, \textit{a.k.a}, similarities between anchors and corresponding positive samples. The related distribution of negative samples also is calculated, because the false negative is annotated as negative. We therefore obtain the positive and negative similarities as:
\begin{equation}
\begin{aligned}
        S^+ &= [s_1^+, s_2^+, s_3^+, \cdots, s_i^+, \cdots], \\ S^- &= [s_1^-, s_2^-, s_3^-, \cdots, s_i^-, \cdots] ,
\end{aligned}
\end{equation}
where $S^+$ and $S^-$ are defined as the sets of matched pairs similarity and mismatched pairs similarity, respectively. Following~\cite{zhang2022negative}, we utilize normal distribution to model the similarities as:
\begin{equation}
\begin{aligned}
        f_{S|c}(s)=\frac{1}{\sigma^{+} \sqrt{2 \pi}} e^{\left[-\frac{\left(s-\mu^{+}\right)^2}{2\left(\sigma^{+}\right)^2}\right]}, \\ f_{S|\bar{c}}(s)=\frac{1}{\sigma^{-} \sqrt{2 \pi}} e^{\left[-\frac{\left(s-\mu^{-}\right)^2}{2\left(\sigma^{-}\right)^2}\right]},
\end{aligned}
\label{eq:eq_5}
\end{equation}
where $(\mu^+, \sigma^+)$ and $(\mu^-, \sigma^-)$ are the mean and standard deviation of the two distributions respectively, which are accumulated and calculated in each mini-batch. $c$ and $\bar{c}$ are random events of match and non-match, respectively.
Note that we have not constructed the distributions leveraging off-the-shelf models or features offlinely, because the representation ability of the model is growing with the training process. In other words, fixed distributions cannot precisely describe the statistical property of similarity for the purpose of eliminating the false negatives for current model. The off-the-shelf model cannot provide suitable hard negative and false negative measurement when the ability of current model surpassed it, and finally limits to further improve the ability of current model. However, due to the similarity being calculated by the current model, there is a certain error. Therefore, in order to better approximate the true similarity distribution of matching and non-matching pairs, only when the similarity of matching pairs in the current mini-batch is higher than that of other non-matching pairs, their similarity will be sampled for calculating the mean and variance to build the distribution.

After obtaining the prior similarity distributions, the posterior probability indicating the likelihood of given negative sample to be a false negative could be expressed as a conditional probability as $P({C=c}| {S=s})$, where $c$ indicates the event of a given sample matches with the anchor in semantics, and $s$ is the similarity score between them. However, since $C$ is a discrete random variable, while $S$ is a continuous one, the conditional probability $P({C=c}| {S=s})$ cannot be directly calculated because the event $\{S=s\}$ with zero probability for continuous random variable. Following~\cite{bertsekas2008introduction}, instead of conditioning on the event $\{S=s\}$, we try to condition on the event $\{{s}\leq {S} \leq {s}+\triangle {s}\}$, where $\triangle {s}$ is a small positive number, and then take the limit as $\triangle {s}$ tends to zero. The posterior probability $P({C=c}| {S=s})$ can be derived following the Bayes' rule:
%\begin{small}
\begin{equation}
\begin{aligned}
    {P({C=c}|S=s)} &\approx {P({C=c}|{s}\leq {S} \leq {s}+\triangle {s})} \\
    &=\frac{P(c)P({s}\leq {S} \leq {s}+\triangle {s}|{c}) }{P({s}\leq {S} \leq {s}+\triangle {s})} \\
\end{aligned}
\end{equation}
%\end{small}
where $s$ represents the similarity between given sample and anchor. The mean value theorem of integrals is subsequently applied as: 
\begin{equation}
\begin{aligned}
   \frac{P(c)P({s}\leq {S} \leq {s}+\triangle {s}|{c}) }{P({s}\leq {S} \leq {s}+\triangle {s})} &=\frac{P(c) \int_{s}^{s+ \triangle s}f_{S|c}(t) dt}{\int_{s}^{s+ \triangle s}f_{S}(t) dt}\\
    &\approx \frac{P(c)f_{S|c}(s)\triangle{s}}{f_{S}(s)\triangle{s}} \\
    &= \frac{P(c)f_{S|c}(s)}{f_{S}(s)} .
\end{aligned}
\label{eq:eq_7}
\end{equation}
Based on the total probability theorem, the denominator $f_{S}(s)$ can be evaluated as:
\begin{equation}
    f_{S}(s) = P(c)f_{S|c}(s) + P(\bar{c})f_{S|\bar{c}}(s),
\end{equation}
where $\bar{c}$ represents the event of a given sample mismatches the anchor. Finally, we can re-write Equation~\ref{eq:eq_7} as:
\begin{equation}
    P(C=c|S=s) = \frac{P(c)f_{S|c}(s)}{P(c)f_{S|c}(s) + P(\bar{c})f_{S|\bar{c}}(s)}.
    \label{eq:eq_9}
\end{equation}

From above derivations, the posterior probability $P(C=c|S=s)$ is determined by three distributions. The first two distributions, namely the similarity distributions of matches and mismatches, have been obtained in Equation~\ref{eq:eq_5}. As everyone knows, there only exist \textit{match} and \textit{mismatch} for a given sample and an anchor, which means the random variable $C$ follows the Bernoulli distribution as:
\begin{equation}
\label{eq:eq_10}
    C\sim B(1,p),
\end{equation}
where $p$ is the probability for event $c$. Since there exist false negatives in the dataset, directly calculating the distribution parameter $p$ conditioned on the annotated data is not appropriate. We simply tune $p$ and choose the best one depending on validation set.

Given a negative sample for an anchor, we can obtain the probability of how likely it would be a false negative from Equation~\ref{eq:eq_9}. Now we need to decide whether the negative should be included in the triplet loss for optimization with the derived false negative probability. Threshold value method would be a straightforward and simple way but a little bit tricky, which needs exhaust tuning for the threshold probability as a hyper-parameter. We therefore follow another intuitive idea that the negative samples with higher probability to be a false negative should occur less times in the triplet, and implement weighted sampling based on the posterior probability. To make the sampling more smooth, we employ the exponential activation to the sampling function, and the whole process can be formulated as:
\begin{equation}
   p_i = exp{(-P(C=c|S=s(d_{i}^{-})))},
\end{equation}
where $p_i$ represents the sampling weight, which will be used for selecting the negative samples. $d_{i}^{-}$ denotes the negative sample.

Adopting the sampling method mentioned above can reduce the occurrence possibility of false negatives. However, to make the optimization object focus on hard negatives, we introduce a cut-down strategy for the easy negatives, whose false negative probability tends to zero and should not be utilized the FNE sampling strategy (as the gray area in the upper right corner of Figure~\ref{fig:overview}). This is because these negative samples are obviously irrelevant to the anchor, and definitely do not match the anchor. As pointed out in~\cite{xuan2020hard,zhang2022negative,chen2020adaptive}, including these easy negatives in the triplet would not provide any useful information for model optimization. Therefore, for negative samples whose false negative probability tends to be zero, we cut down their sampling weights and reset them as follows:
\begin{equation}
\label{eqution:uninfor}
   p_i = exp{ (-\alpha(s(d_{i}^{-}) - s(d^{+}))^{2}) },
\end{equation}
where $\alpha$ is a hyper-parameter to control the density of the sampling. Finally, the negative sampling for the False Negative Elimination strategy is a combination of these two, as shown below:
\begin{equation}
\begin{aligned}
\label{eq:eq_13}
        p_i = \begin{cases}
        exp{(-P(C=c|S=s(d_{i}^{-})))}, &others \\
        exp{ (-\alpha(s(d_{i}^{-}) - s(d^{+}))^{2}) }, &P(C=c|S=s(d_{i}^{-})) \leq \lambda
        \end{cases},
\end{aligned}
\end{equation}
where $\lambda$ is set as 0.01. In this way, false negatives and easy negatives are penalized with smaller sampling weights, making the model learn semantic representations based on true and hard negatives.

%%%%%%%%%%%%%%%%%==Table Block==%%%%%%%%%%%%%%%
\begin{table*}
\centering
\caption{Performance comparison with baselines on Flick30K and MS-COCO 1K test set. The best results are shown in bold.}
\begin{tabular}{c|cccccc|cccccc}
\hline
\multirow{3}{*}{Methods} & \multicolumn{6}{c|}{Flickr30K}                                                  & \multicolumn{6}{c}{MS-COCO}                                               \\ \cline{2-13} 
                        & \multicolumn{3}{c|}{Image-to-Text}      & \multicolumn{3}{c|}{Text-to-Image} & \multicolumn{3}{c|}{Image-to-Text}      & \multicolumn{3}{c}{Text-to-Image} \\ \cline{2-13} 
                        & R@1  & R@5  & \multicolumn{1}{c|}{R@10} & R@1        & R@5       & R@10      & R@1  & R@5  & \multicolumn{1}{c|}{R@10} & R@1       & R@5       & R@10      \\ \hline
SCAN$_\textit{(ECCV'18)}$~\cite{lee2018stacked}                    & 67.4 & 90.3 & \multicolumn{1}{c|}{95.8} & 48.6       & 77.7      & 85.2      & 72.7 & 94.8 & \multicolumn{1}{c|}{98.4} & 58.8      & 88.4      & 94.8      \\
CAMP$_\textit{(CVPR'19)}$~\cite{wang2019camp}                    & 68.1 & 89.7 & \multicolumn{1}{c|}{95.2} & 51.5       & 77.1      & 85.2      & 72.3 & 94.8 & \multicolumn{1}{c|}{98.3} & 58.5      & 87.9      & 95.0      \\
BFAN$_\textit{(MM'19)}$~\cite{liu2019focus}                    & 68.1 & 91.4 & \multicolumn{1}{c|}{-}    & 50.8       & 78.4      & -         & 74.9 & 95.2 & \multicolumn{1}{c|}{-}    & 59.4      & 88.4      & -         \\
SAEM$_\textit{(MM'19)}$~\cite{wu2019learning}                    & 69.1 & 91.0 & \multicolumn{1}{c|}{95.1} & 52.4       & 81.1      & 88.1      & 71.2 & 94.1 & \multicolumn{1}{c|}{97.7} & 57.8      & 88.6      & 94.9      \\
%DP-RNN~\cite{chen2020expressing}$_\textit{(AAAI'20)}$                  & 70.2 & 91.6 & \multicolumn{1}{c|}{95.8} & 55.5       & 81.3      & 88.2      & 75.3 & 95.8 & \multicolumn{1}{c|}{98.6} & 62.5      & 89.7      & 95.1      \\
VSRN$_\textit{(ICCV'19)}$~\cite{li2019visual}                    & 71.3 & 90.6 & \multicolumn{1}{c|}{96.0} & 54.7       & 81.8      & 88.2      & 76.2 & 94.8 & \multicolumn{1}{c|}{98.2} & 62.8      & 89.7      & 95.1      \\
CAAN$_\textit{(CVPR'20)}$~\cite{zhang2020context}                    & 70.1 & 91.6 & \multicolumn{1}{c|}{97.2} & 52.8       & 79.0      & 87.9      & 75.5 & 95.4 & \multicolumn{1}{c|}{98.5} & 61.3      & 89.7      & 95.2      \\
%SGM~\cite{wang2020cross}$_\textit{(WACV'20)}$                     & 71.8 & 91.7 & \multicolumn{1}{c|}{95.5} & 53.5       & 79.6      & 86.5      & 73.4 & 93.8 & \multicolumn{1}{c|}{97.8} & 57.5      & 87.3      & 94.3      \\
IMRAM$_\textit{(CVPR'20)}$~\cite{chen2020imram}                   & 74.1 & 93.0 & \multicolumn{1}{c|}{96.6} & 53.9       & 79.4      & 87.2      & 76.7 & 95.6 & \multicolumn{1}{c|}{98.5} & 61.7      & 89.1      & 95.0      \\
MMCA$_\textit{(CVPR'20)}$~\cite{wei2020multi}                    & 74.2 & 92.8 & \multicolumn{1}{c|}{96.4} & 54.8       & 81.4      & 87.8      & 74.8 & 95.6 & \multicolumn{1}{c|}{97.7} & 61.6      & 89.8      & 95.2      \\
GSMN$_\textit{(CVPR'20)}$~\cite{liu2020graph}                    & 76.4 & 94.3 & \multicolumn{1}{c|}{97.3} & 57.4       & 82.3      & 89.0      & 78.4 & 96.4 & \multicolumn{1}{c|}{98.6} & 63.3      & 90.1      & 95.7      \\
ADAPT$_\textit{(ECCV'20)}$~\cite{chen2020adaptive}                   & 76.6 & 95.4 & \multicolumn{1}{c|}{97.6} & 60.7       & 86.6      & 92.0      & 76.5 & 95.6 & \multicolumn{1}{c|}{98.9} & 62.2      & 90.5      & 96.0      \\
CAMERA$_\textit{(MM'20)}$~\cite{qu2020context}                  & 78.0 & 95.1 & \multicolumn{1}{c|}{97.9} & 60.3       & 85.9      & 91.7      & 77.5 & 96.3 & \multicolumn{1}{c|}{98.8} & 63.4      & 90.9      & 95.8      \\
DIME$_\textit{(SIGIR'21)}$~\cite{qu2021dynamic}                    & 81.0 & 95.9 & \multicolumn{1}{c|}{98.4} & 63.6       & 88.1      & 93.0      & 78.8 & 96.3 & \multicolumn{1}{c|}{98.7} & 64.8      & 91.5      & 96.5      \\
VSE$\infty$$_\textit{(CVPR'21)}$~\cite{chen2021learning}             & 81.7 & 95.4 & \multicolumn{1}{c|}{97.6} & 61.4       & 85.9      & 91.5      & 79.7 & 96.4 & \multicolumn{1}{c|}{98.9} & 64.8      & 91.4      & 96.3      \\
MV-VSE$_\textit{(IJCAI'22)}$~\cite{li2022multi}                    & 82.1 & 95.8 & \multicolumn{1}{c|}{97.9} & 63.1       & 86.7      & 92.3      & 80.4 & 96.6 & \multicolumn{1}{c|}{99.0} & 64.9      & 91.2      & 96.0      \\
NAAF$_\textit{(CVPR'22)}$~\cite{zhang2022negative}                    & 81.9 & 96.1 & \multicolumn{1}{c|}{98.3} & 61.0       & 85.3      & 90.6      & 80.5 & 96.5 & \multicolumn{1}{c|}{98.8} & 64.1      & 90.7      & 96.5      \\
CMSEI$_\textit{(WACV'23)}$~\cite{ge2023cross}                    & 82.3 & 96.4 & \multicolumn{1}{c|}{98.6} & 64.1       & 87.3      & 92.6      & 81.4 & 96.6 & \multicolumn{1}{c|}{98.8} & 65.8      & 91.8      & \textbf{96.8}      \\
RVSE++$_\textit{(Arxiv'23)}$~\cite{li2023selectively}                  & 83.6 & 96.5 & \multicolumn{1}{c|}{98.6} & 64.3       & 88.2      & 93.0      & 81.6 & 96.6 & \multicolumn{1}{c|}{98.8} & 66.6      & \textbf{92.1}      & {96.6}      \\ \hline
Ours                     & \textbf{85.4} & \textbf{98.1} & \multicolumn{1}{c|}{\textbf{99.2}} & \textbf{70.1}       & \textbf{90.7}      & \textbf{94.7}      & \textbf{82.5} & \textbf{96.6} & \multicolumn{1}{c|}{\textbf{99.0}} & \textbf{67.7}      & 90.5      & 95.2      \\ \hline
\end{tabular}
\label{tab:tab1}
\vspace{-5pt}
\end{table*}
%%%%%%%%%%%%%%%%%==Table Block==%%%%%%%%%%%%%%%

\subsection{Momentum Memory Module}
\label{sec:sec3_3} 
Limited by the GPU memory size, many image-text matching methods~\cite{lee2018stacked, faghri2017vse++, zhang2022negative, qu2021dynamic} implement very small batch size, \myeg{}, 32 or 64, for training based on large models, \myeg{}, BERT. Within such small mini-batch, there only exists small number of negatives for each anchor, most of which are easy negatives and redundant to learn the distance between the anchor and positives~\cite{xuan2020hard}. To make more hard negatives and false negative accessible for each iteration, we devise a novel momentum memory module, consisting of an image memory bank and a text memory bank, to construct a large buffer and dynamically retain more negatives for sampling.

\noindent \textbf{Image Memory Bank.} To mine negative samples for text queries, we construct an image memory bank $\mathbf{M_v} \in \mathbb{R}^{K \times d}$ to store image representations, which is intrinsically a queue of features. During each iteration, the image representations in current mini-batch will be enqueued into $\mathbf{M_v}$, and the oldest images in the buffer queue will be removed due to the fixed size of the queue. Finally, all image representations in the memory bank will be used to calculate the similarity and loss with the text representations of the current mini-batch. Since the size of the memory bank would be very large to access more negatives during sampling, the representations in the same memory bank come from different mini-batches extracted by model parameters across multiple iterations, which exist feature shifts and raise issues to the similarity ranking. Inspired by~\cite{he2020momentum, zhang2023user}, we incorporate a momentum encoder (shown on the left in Figure~\ref{fig:overview}) with our image memory bank to alleviate this issue. Different from the original image encoder $\theta_q^v$ that is directly updated through the gradient back-propagation, the momentum encoder $\theta_k^v$ keeps a more smooth update with momentum from previous steps. The update process can be formulated as:
\begin{equation}
\theta_k^v \leftarrow m\theta_k^v + (1-m)\theta_q^v,
\end{equation}
where $m \in [0,\ 1)$ is a momentum hyper-parameter, we set 0.995 in this paper.
With the momentum technique, the feature shifts could be significantly alleviated and make the features in the same memory bank more suitable for similarity calculation. 

\noindent \textbf{Text Memory Bank.} Similarly, we also construct a text memory bank $\mathbf{M_t} \in \mathbb{R}^{K \times d}$ with momentum update scheme for text representations storage. $K$ is the length of memory bank. 

\subsection{Training and Inference}
\label{sec:sec3_4} 

\noindent \textbf{Training Objective.} Following the existing methods~\cite{faghri2017vse++, lee2018stacked, chen2021learning, zhang2022negative,wang2022point}, we also implement the triplet loss in Equation~\ref{equ:triplet} as our loss function. Different from previous works, however, the negative samples selected for repulsion are no longer the hard negative samples, but the ones sampled via our false negative elimination strategy. The final loss function can be formulated as follows:
\begin{equation}
\label{equ:triplet_final}
\begin{aligned}
        \mathbf{\mathcal{L}} = &[margin - s(\mathbf{\bar{v}}, \mathbf{\bar{w}}) + s(\mathbf{\bar{v}}, \mathbf{\bar{w}_{FNE}})]_+ \\ + \ &[margin - s(\mathbf{\bar{v}}, \mathbf{\bar{w}}) + s(\mathbf{\bar{v}_{FNE}}, \mathbf{\bar{w}})]_+ ,
\end{aligned}
\end{equation}
where $\mathbf{\bar{w}_{FNE}}$ and $\mathbf{\bar{v}_{FNE}}$ denote the negative embedding for text and image respectively, selected by our FNE strategy.

\noindent \textbf{Inference Process.} During the inference stage, the negatives sampling strategy, in specific our FNE strategy, will not be implemented any more. Our framework follows the commonly-used way that computes the similarity between the query anchor and all the samples in the test set, and then ranks them with similarities. The samples with highest rank, \myeg{}, top-1, top-5, and top-10, would be considered as the retrieval results. 

\section{Experiments}

We evaluate our method on Flickr30K~\cite{young2014image} and MS-COCO~\cite{chen2015microsoft}. The datasets and experimental details are presented in the appendix.

\subsection{Performance Comparison}

\noindent \textbf{Baselines and state-of-the-arts.} To evaluate the effectiveness of our proposed FNE strategy, we compare it with several state-of-the-art baselines on the MS-COCO and Flickr30K datasets in Table~\ref{tab:tab1} and Table~~\ref{tab:tab2}. As summarized in Section~\ref{sec:rel_1}, these baselines include (1) the Visual Semantic Embedding methods, including SAEM~\cite{wu2019learning}, CAMERA\cite{qu2020context}, VSE$\infty$~\cite{chen2021learning}, MV-VSE~\cite{li2022multi}, \myec{}. (2) the Cross-Attention methods, including SCAN~\cite{lee2018stacked}, IMRAM~\cite{chen2020imram}, DIME~\cite{qu2021dynamic}, NAAF\cite{zhang2022negative}, \myec{}.  
The former fashion, VSE method, is efficient in retrieval during inference, but its performance is generally limited due to the lack of sufficient inter-modal interactions. The latter one, CA fashion explores the fine-grained cross-modal interactions and achieves better performance, while it suffers from expensive computation cost. 
However, due to the rise of pre-trained models such as BERT~\cite{devlin2018bert}, the performance gap between the two types of methods is gradually being eliminated. The recent state-of-the-art method, RVSE++, is precisely a visual semantic embedding method. Note that, many methods~\cite{qu2021dynamic, qu2020context, chen2020adaptive, lee2018stacked} achieve further improvement through ensemble of multiple models. In order to provide a fair and comprehensive comparison, we also provide an ensemble model, which averages similarity scores of two models with different random seeds.

%%%%%%%%%%%%%%%%%==Table Block==%%%%%%%%%%%%%%%
\begin{table}[]
\caption{Performance comparison with state-of-the-art baselines on MS-COCO 5K test set.}
\begin{tabular}{c|ccc|ccc}
\hline
\multirow{2}{*}{Methods} & \multicolumn{3}{c|}{Image-to-Text} & \multicolumn{3}{c}{Text-to-Image} \\ \cline{2-7} 
                        & R@1        & R@5       & R@10      & R@1       & R@5       & R@10      \\ \hline
%SCAN(single)            & 46.4       & 77.4      & 87.2      & 34.4      & 63.7      & 75.7      \\
SCAN~\cite{lee2018stacked}                    & 50.4       & 82.2      & 90.0      & 38.6      & 69.3      & 80.4      \\
CAMP~\cite{wang2019camp}                    & 50.1       & 82.1      & 89.7      & 39.0      & 68.9      & 80.2      \\
VSRN~\cite{li2019visual}                    & 53.0       & 81.1      & 89.4      & 40.5      & 70.6      & 81.1      \\
CAAN~\cite{zhang2020context}                    & 52.5       & 83.3      & 90.9      & 41.2      & 70.3      & 82.9      \\
%SGM~\cite{wang2020cross}                     & 50.0       & 79.3      & 87.9      & 35.3      & 64.9      & 76.5      \\
IMRAM~\cite{chen2020imram}                   & 53.7       & 83.2      & 91.0      & 39.6      & 69.1      & 79.8      \\
MMCA~\cite{wei2020multi}                    & 54.0       & 82.5      & 90.7      & 38.7      & 69.7      & 80.8      \\
%CAMERA(single)          & 53.1       & 81.3      & 89.8      & 39.0      & 70.5      & 81.5      \\
CAMERA~\cite{qu2020context}                  & 55.1       & 82.9      & 91.2      & 40.5      & 71.7      & 82.5      \\
%DIME(single)            & 56.1       & 83.2      & 91.1      & 40.2      & 70.7      & 81.4      \\
DIME~\cite{qu2021dynamic}                    & 59.3       & 85.4      & 91.9      & 43.1      & 73.0      & 83.1      \\ 
VSE$\infty$~\cite{chen2021learning}             & 58.3       & 85.3      & 92.3      & 42.4      & 72.7      & 83.2      \\
MV-VSE~\cite{li2022multi}                    & 59.1       & 86.3      & 92.5      & 42.5      & 72.8      & 83.1      \\
NAAF~\cite{zhang2022negative}                    & 58.9       & 85.2      & 92.0      & 42.5      & 70.9      & 81.4      \\
CMSEI~\cite{ge2023cross}                    & 61.5       & 86.3      & 92.7      & 44.0      & 73.4      & 83.4      \\
RVSE++~\cite{li2023selectively}                  & 60.6       & 86.4      & 92.8      & 44.5      & 74.5      & \textbf{84.5}      \\
\hline
%FNAM(single)             & {}       & {}      & {}      & {}      & {}      & {}      \\
Ours                     & \textbf{64.3}       & \textbf{87.0}      & \textbf{92.9}      & \textbf{47.6}      & \textbf{74.9}      & {83.5}      \\ \hline
\end{tabular}
\label{tab:tab2}
\end{table}
%%%%%%%%%%%%%%%%%==Table Block==%%%%%%%%%%%%%%%

\noindent \textbf{Performance comparison.} The quantitative result comparisons between our FNE and SOTAs are shown in Table~\ref{tab:tab1} and Table~\ref{tab:tab2}. Firstly, we can observe that the NAAF and RVSE++ methods outperform other baseline methods in the CA and VSE methods, respectively. NAAF is an extension of SCAN~\cite{lee2018stacked} that further refines the similarity calculation process by focusing on the similarity of mismatched fragments. This allows the CA methods to consider both positive and negative effects during the attention calculation process, resulting in better CA performance. On the other hand, RVSE++, based on VSE$\infty$~\cite{chen2021learning}, chooses whether to mine hard negative samples according to the similarity between positive and negative pairs, thus alleviating the gradient vanishing problem and achieving superior performance. This also verifies that hard negative mining is crucial for image-text matching, and RVSE++ improves performance by mining hard negative samples at the appropriate time. Our FNE steps further to investigate the false negative issue in hard negative mining, and outperforms almost all the state-of-the-art methods across most evaluation metrics, except for R@5 and R@10 for the task of text-to-image matching on MS-COCO. 
Specifically, on the Flickr30K dataset, compared with the RVSE++, our FNE achieves 1.8\% and 5.8\% improvements in terms of R@1 at two retrieval directions, respectively. Similarly, compared with NAAF, our FNE achieves 3.5\% and 9.1\% improvements in terms of R@1 at two directions, respectively. We also achieve similar gains on the larger and complicated MS-COCO dataset, including the evaluation metrics for the 5-folds 1k and full 5k. Compared with the state-of-the-art model RVSE++ on MS-COCO 1K test set, our FNE achieves 0.9\% and 1.1\% improvements in terms of R@1 at two retrieval directions, respectively. Moreover, on the MS-COCO Full 5K test set (shown in Table~\ref{tab:tab2}), our FNE also achieves 3.7\% and 3.1\% improvements in terms of R@1 at two directions, respectively.
All experimental results demonstrate the effectiveness of our FNE strategy. Our approach effectively alleviates the prevalent false negative problem in image-text matching systems, enabling the model to learn more accurate and discriminative representations and resulting in better image-text matching performance.

%%%%%%%%%%%%%%%%%==Table Block==%%%%%%%%%%%%%%%
\begin{table}[]
\centering
\caption{Ablation studies about the each component of model, which are obtained on Flickr30K.} 
% \resizebox{\linewidth}{!}{
\setlength{\tabcolsep}{1.5mm}{
\begin{tabular}{c|ccc|ccc}
\hline
\multirow{2}{*}{Methods} & \multicolumn{3}{c|}{Image-to-Text} & \multicolumn{3}{c}{Text-to-Image} \\ \cline{2-7} 
                        & R@1        & R@5       & R@10      & R@1       & R@5       & R@10      \\ \hline
w/o MMM+FNE                 & 77.2       & 95.2      & 98.6      & 63.7      & 88.3      & 93.0      \\
w/o MMM                  & 77.0        & 95.0       & 98.3       & 63.2       & 89.0       & 93.9       \\
w/o FNE                 & 83.8       & 97.0      & 99.1      & 67.0      & 88.8      & 93.1      \\ \hline
Full                     & 84.7       & 97.1      & 99.1      & 68.1      & 90.0      & 94.0      \\ \hline
\end{tabular}
}
\label{tab:abalation}
\end{table}
%%%%%%%%%%%%%%%%%==Table Block==%%%%%%%%%%%%%%%

%%%%%%%%%%%%%%%%%==Table Block==%%%%%%%%%%%%%%%
\begin{table}[]
\centering
\caption{Investigation of the matching probability $p$ in Equation~\ref{eq:eq_10}. Experiments are conducted on Flickr30K.} 
% \resizebox{\linewidth}{!}{
\setlength{\tabcolsep}{1.5mm}{
\begin{tabular}{c|ccc|ccc}
\hline
\multirow{2}{*}{Methods} & \multicolumn{3}{c|}{Image-to-Text} & \multicolumn{3}{c}{Text-to-Image} \\ \cline{2-7} 
                        & R@1        & R@5       & R@10      & R@1       & R@5       & R@10      \\ \hline
%1/1000_no                 & 83.3       & 97.2      & 98.7      & 67.8      & 89.9      & 93.5      \\
1/1000                  & 83.2        & 97.8       & 99.2       & 67.8       & 89.9       & 93.9       \\
%1/5000_no                 & 83.6       & 97.2      & 99.1      & 67.7      & 89.4      & 93.8   \\
1/5000                 & 83.8       & 97.1      & 98.8      & 68.0      & 90.1      & 94.0  \\
%1/10000_no                 & 84.5       & 97.1      & 98.8      & 68.0      & 90.0      & 93.7  \\
1/10000                     & 84.7       & 97.1      & 99.1      & 68.1      & 90.0      & 94.0      \\
1/50000                     & 84.2       & 97.2      & 98.9      & 67.8      & 89.9      & 93.6 \\
1/100000                     & 83.8       & 97.3      & 98.7      & 68.2      & 89.8      & 93.7 \\ \hline
\end{tabular}
}
\label{tab:match_p}
% \vspace{-5pt}
\end{table}
%%%%%%%%%%%%%%%%%==Table Block==%%%%%%%%%%%%%%%

%%%%%%%%%%%%%%%%%==Figure Block==%%%%%%%%%%%%%%%
\begin{figure*}
\begin{center}
\includegraphics[width=0.9\textwidth]{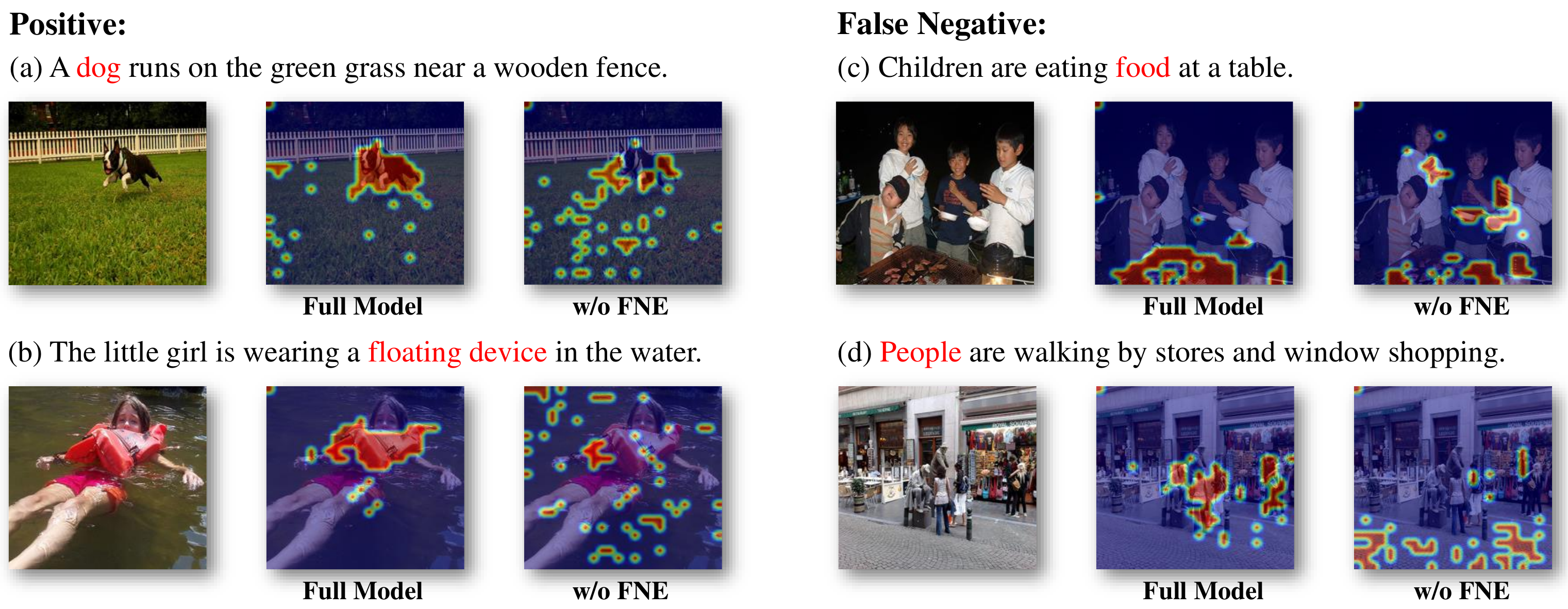}
%7.16
\end{center}
\vspace{-5pt}
\caption{Illustration of cross-attention maps. The left side corresponds to positive samples, while the right side corresponds to false negatives. We compare the Full model and the w/o FNE model.
}
\vspace{-5pt}
\label{fig:hot}
\end{figure*}
%%%%%%%%%%%%%%%%%==Figure Block==%%%%%%%%%%%%%%%

\subsection{Ablation Study}
We perform extensive ablation studies on Flickr30K to evaluate the effectiveness of each component of the proposed approach, and illustrate the results in Table~\ref{tab:abalation} (all the results are with single model).
From the results, we have following observations: 1) The basic model with ViT and BERT only, removing both the false negative elimination strategy and the momentum memory module (w/o MMM+FNE), leads to dramatically performance degradation, which verifies the effectiveness of the combination of proposed FNE and MMM. 2) When we only remove the momentum memory module (w/o MMM), the performance further decreases a little. This is because MMM enlarges the negative pool and makes more negatives accessible in sampling, which could provide more meaningful negatives. 3) When we independently employ the momentum memory module without FNE (w/o FNE), the performance achieve significant improvement comparing with w/o MMM, but there remains a large gap between the full model. This first verifies the conclusion again, that MMM enables more negative in sampling and results in better performance. Meanwhile, it also demonstrates that the false negative elimination strategy is crucial for mining suitable hard negatives for model optimization, which could further improve the image-text matching accuracy.

% \vspace{-10pt}
Besides, we also investigate the effects of setting different values for the prior matching probability $p$ in Equation~\ref{eq:eq_10}. Since the matching examples are rare in real-world scenario, we set a very small number for $p$, \myeg{}, 1/10000, and tune it based on the validation set. The results are shown in Table~\ref{tab:match_p}. It can be observed that the impact of different parameter settings on the final retrieval performance is not very significant, even with a hundredfold difference between 1/1000 and 1/100000. This indicates that the exponential function we employed when generating the final weights effectively reduces the sensitivity of our method to this parameter, resulting in a more robust retrieval performance. We observe that the model achieves the best performance when $p=1/10000$, and therefore fix the value for all the experiments.

%%%%%%%%%%%%%%%%%==Figure Block==%%%%%%%%%%%%%%%
\begin{figure}
    \centering
    \includegraphics[width = \linewidth]{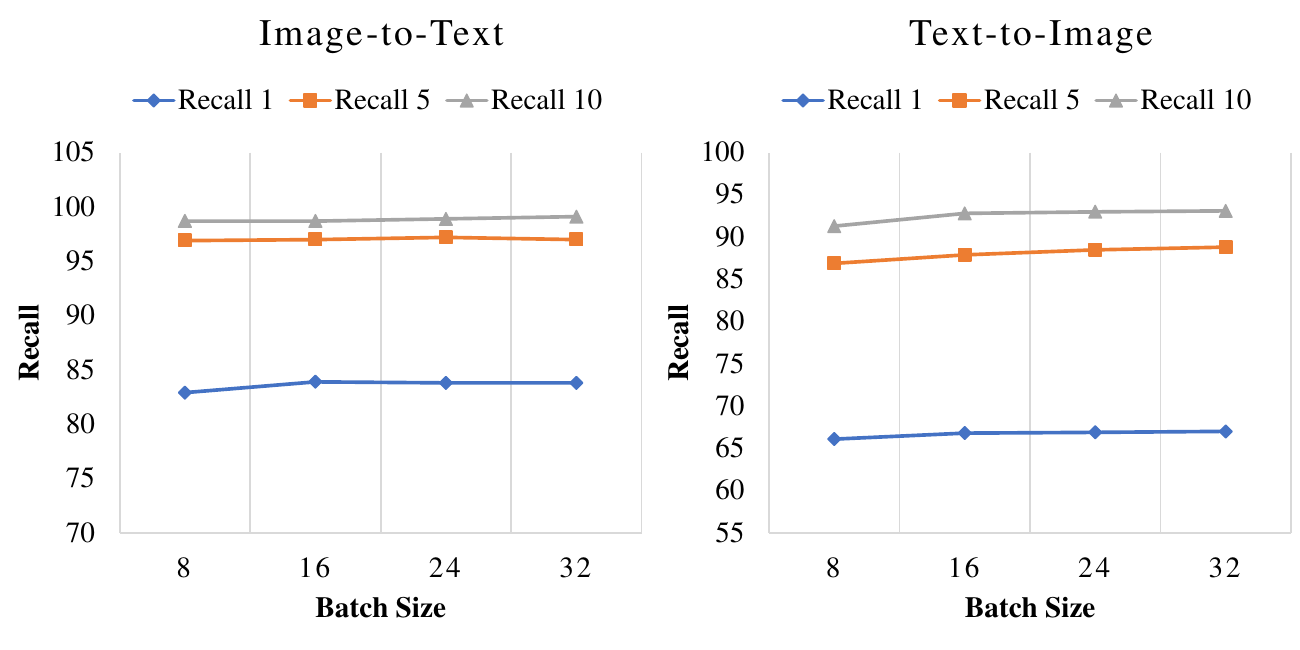}
    \caption{Effects of different mini-batch sizes on the momentum memory module with fixed length.}
    \label{fig:batch_size}
    \vspace{-10pt}
\end{figure}
%%%%%%%%%%%%%%%%%==Figure Block==%%%%%%%%%%%%%%%

\subsection{Analysis of Memory Module}
Previous works~\cite{chen2020adaptive, chen2017beyond, zhang2022unified} point out that larger mini-batch usually leads to better image-text matching results due to accessing of more negative samples. 
As aforementioned, the memory module enlarges the sampling pool and has addressed the problem of less negative samples with small batch-size. To further investigate the effectiveness for more small batch, we conduct experiments with fixed memory length, \myeg{}, 8192, across different small batch-sizes, \myeg{}, 24 and 16, even 8. The experimental results are illustrated in Figure~\ref{fig:batch_size}, from which we can observe that the model with different batch size exhibits similar performance, which means it is no longer sensitive to the batch size with our large memory module. This makes our approach able to perform on the devices with small memory with small batch size, \myeg{}, 8 samples in a mini-batch.

% \vspace{-5pt}
\subsection{Qualitative Analysis and Visualization} 
In this paper, we have defined the false negatives as the samples sharing the same semantics with anchor. Obviously, pushing away such false negatives from the anchor would confuse the model in metric learning and lead to inferior semantic representations. To examine the concrete effects on representation learning, we visualize the attention map in images for specific words, as illustrated in Figure~\ref{fig:hot}, which indicates the fine-grained associations between images and texts. It can be observed that our FNE produces features that more accurately capture objects with respect to the target word, both on positive samples and false negatives. The attention maps produced by the w/o FNE model are scattered and inaccurate enough. 
To be noted that, when we remove FNE strategy, the false negative will be pushed away as a negative, and learn to associates the word to the background (as shown on the right side). By excluding the false negatives in the triplet, the model learns more accurate and distinctive feature representations, and results in better image-text matching performance. We also illustrate more examples in the supplementary, w.r.t. the effects on representation learning and sampling probability of FNE. Please kindly find them.
\vspace{-5pt}

\section{Conclusions}
In this paper, we investigated the problem of false negative in hard negative mining for image-text matching. To alleviate this issue, we proposed a novel False Negative Elimination (FNE) strategy, which first calculated the false negative probability of each negative based on Bayes' rule, and implemented weighted-sampling with the probability to decide whether a negative sample to be included in the triplet loss. To make more negative samples accessible during the sampling process, we further introduced a momentum memory module to enlarge the sampling pool of negatives. With the above techniques, the false negatives could be well eliminated and made the model focus on the learning of true and hard negatives.
Extensive experiments had been conducted on Flickr30K and MS-COCO, and the results verified the effectiveness of the proposed approach.

\begin{acks}
This work was partially supported by the National Natural Science Foundation of China
under grant 62220106008, U20B2063, and 62102070, and partially supported by Sichuan Science and Technology Program under grant 2023NSFSC1392.
This research was also supported by the National Research Foundation, Singapore under its Industry Alignment Fund – Pre-positioning (IAF-PP) Funding Initiative. Any opinions, findings and conclusions or recommendations expressed in this material are those of the author(s) and do not reflect the views of National Research Foundation, Singapore.
\end{acks}

\bibliographystyle{ACM-Reference-Format}
\bibliography{fne}


\appendix

\section{Appendix}

%%%%%%%%%%%%%%%%%==Figure Block==%%%%%%%%%%%%%%%
\begin{figure*}[ht]
    \begin{center}
    \includegraphics[width = 0.8\linewidth]{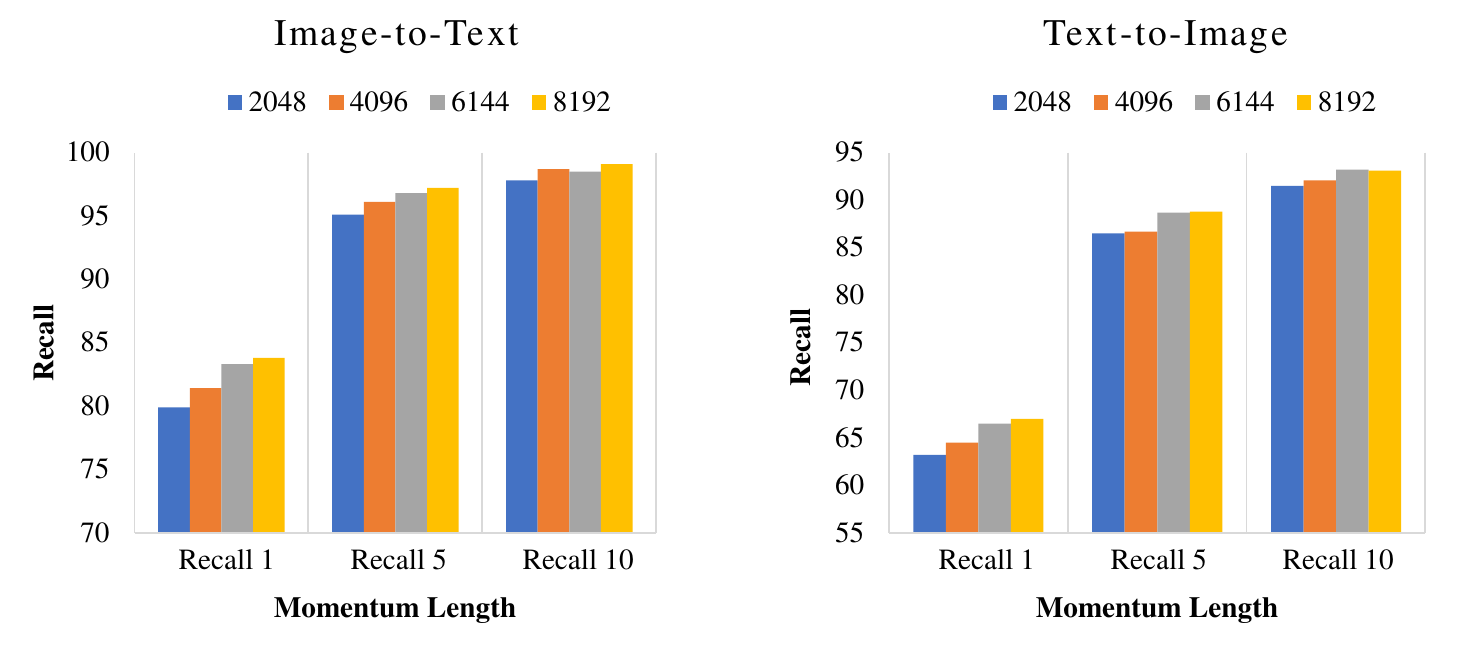}
    \end{center}
    \caption{Effects of momentum memory module with different length.}
    \label{fig:mmm_size}
\end{figure*}
%%%%%%%%%%%%%%%%%==Figure Block==%%%%%%%%%%%%%%%

%%%%%%%%%%%%%%%%%==Figure Block==%%%%%%%%%%%%%%%
\begin{figure*}[ht]
\begin{center}
\includegraphics[width=0.95\textwidth, height=0.7\textwidth]{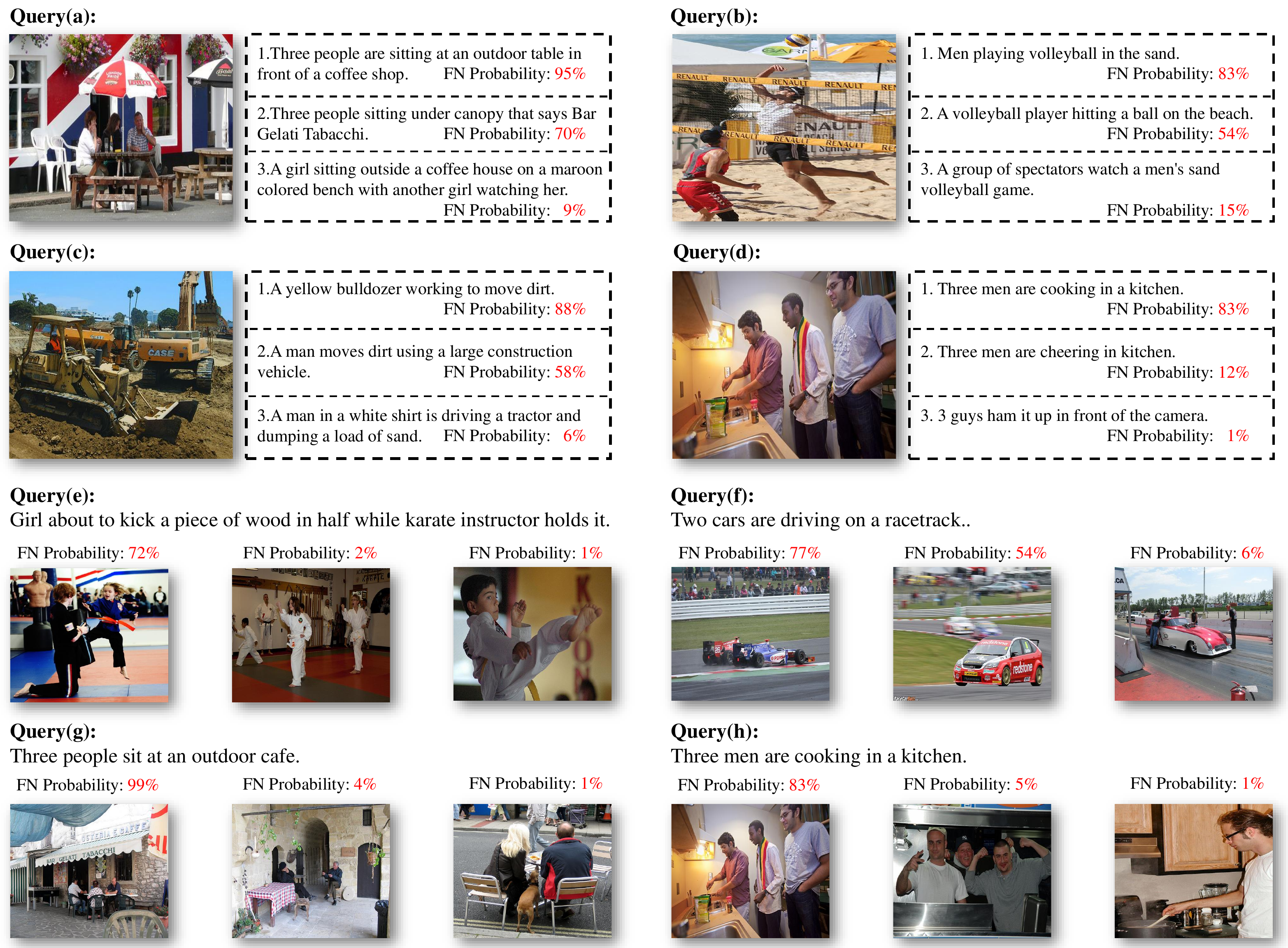}
%7.16
\end{center}
% \vspace{-5pt}
\caption{
Illustration of the estimated false negative probability for \textcolor{red}{negative samples}. FN Probability refers to the false negative probability.
}
% \vspace{-10pt}
\label{fig:p}
\end{figure*}
%%%%%%%%%%%%%%%%%==Figure Block==%%%%%%%%%%%%%%%

Below, we first introduce the details of the dataset and experimental implementation, then investigate the effects of the size of memory module, and present several cases to examine the false negative probability estimated by our strategy. In addition, we also visualize more examples to enrich the analysis described in Section 4.6.

\subsection{Datasets and Evaluation Metrics}

{\bf Flickr30K} comprises 31,783 images, each of which is associated with 5 different sentences. According to the previous settings in~\cite{karpathy2015deep, lee2018stacked, liu2019focus}, we split this dataset into 29,783 training images, 1,000 validation images, and 1,000 testing images. {\bf MS-COCO} dataset comprises 123,287 images, each of which is also annotated with five different sentences. To ensure fair comparison with previous works, we used the Karpathy split~\cite{karpathy2015deep} to divide the dataset into 113,287 images for training, 5,000 images for validation, and 5,000 images for testing. Notably, in the MS-COCO, we conduct tests not only on the full 5K-image test set, but also on the 1K-image test set (averaged over 5 folds).

For the evaluation on Flickr30K and MS-COCO, following~\cite{chen2015microsoft}, we adopt the Recall@K (R@K), with K = {1, 5, 10} as the evaluation metric for the retrieval results. Recall@K measures the percentage of queries for which the ground-truth hit in its top-K ranking list. The higher R@K indicates the better performance.

\subsection{Implement Details}
The experiments and framework of our method are implemented using PyTorch. The size of the mini-batch and momentum memory module is set to 32 and 8192 respectively, with 25 training epochs on both datasets. We employed the Adam optimizer~\cite{kingma2014adam} with an initial learning rate of 0.00002 for model optimization. The learning rate is decayed by 10 times at the 5th and 15th epochs during training. For the pre-trained ViT and BERT models, we use the basic versions, namely "vit-base-patch16" and "bert-base-uncased". The margin in triplet loss function is empirically set as 0.2. The hyper-parameter $\alpha$ in the Equation~\ref{eqution:uninfor}, adjusting the sampling density, is set as 0.5.

\subsection{The Effects of momentum memory module with different length}
As aforementioned, we have demonstrated that when the length of the momentum memory model is fixed, it can achieve similar performance regardless of the mini-batch size, even if it is small, \myeg{}, 8 samples in a mini-batch. This means that with our momentum memory module, the model could onlinely sample hard negatives with small batch-size, and therefore can be implemented on devices with small memory. To further investigate the effects of the size of our momentum memory module, we conduct experiments with a fixed mini-batch size, 32 in specific, across different lengths of memory banks, \myeg{}, 2048, 4096, 6144, and 8192. The experimental results are shown in Figure~\ref{fig:mmm_size}, from which we can observe that as the length of the memory module increases, the image-text matching performance also improves, indicating that increasing the length of momentum memory module achieves performance improvement similar to increasing the mini-batch size. Because with larger momentum memory bank, our FNE is able to access more hard negatives and select the true hard negatives for training, which could achieve similar effects as a larger mini-batch size without increasing the memory burden on GPU.

\subsection{Illustrations of False Negative Probability}
To better understand the effectiveness of our proposed False Negative Elimination (FNE) strategy, we examine the estimated false negative probabilities for \textbf{negative samples}, and illustrate several examples in Figure~\ref{fig:p}. The upper examples set images as anchors, while the bottom ones correspond to text anchor. Each example listed with a false negative probability is defined as a negative sample in the dataset. However, we can observe that some negative samples obviously share the same semantics with corresponding anchor 
 and should have been matched with the anchor. In previous works, these negative samples were selected as hard negatives due to their high similarity with the anchor and will be pushed away with hard negative mining strategies, resulting in conflicting optimization objectives. With the adoption of our FNE strategy, as shown in the figure, we can estimate the posterior probability of a negative sample to be a false negative, termed as false negative probability in this work, and then implement a weighted-sampling scheme to decrease the occurrence in the triplet. We also can observe that the samples with low false negative probability that indeed share different semantics with the anchor. In summary, our FNE strategy could well estimate the likelihood of a negative sample to be a false negative and mine true hard negatives, resulting in better representation learning and matching performance.

%%%%%%%%%%%%%%%%%==Figure Block==%%%%%%%%%%%%%%%
\begin{figure*}[ht]
\begin{center}
\includegraphics[width=0.95\textwidth]{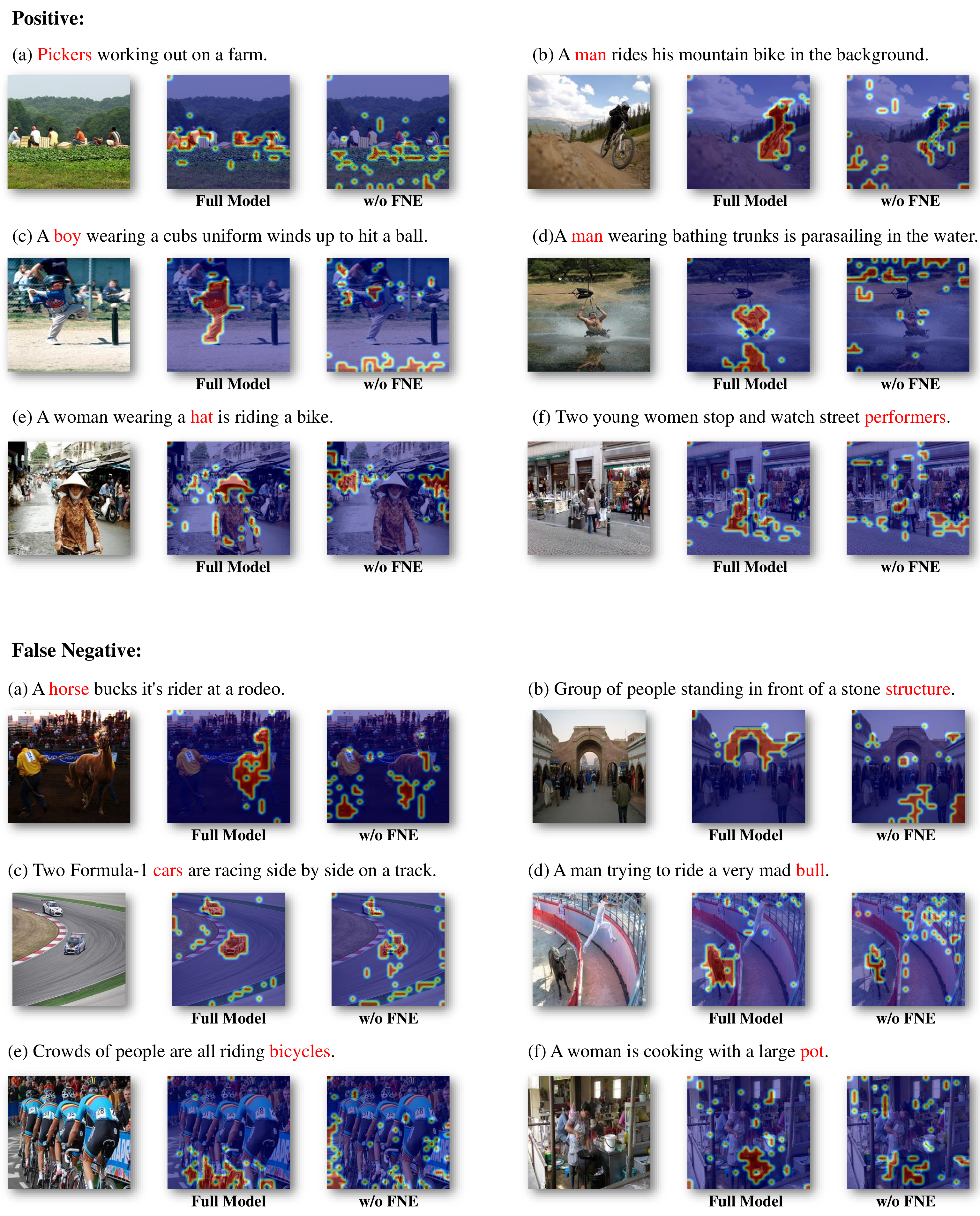}
%7.16
\end{center}
%\vspace{-5pt}
\caption{Illustration of cross-attention maps. The upper side corresponds to positive samples, while the lower side corresponds to false negatives. We compare the Full model and the w/o FNE model.
}
%\vspace{-5pt}
\label{fig:hot}
\end{figure*}
%%%%%%%%%%%%%%%%%==Figure Block==%%%%%%%%%%%%%%%

\subsection{Exhibition of Cross-Attention Maps}
More cross-attention exemplars are presented in Figure~\ref{fig:hot}. It can be observed that our model accurately attends to the target region for both positive samples and false negatives. These examples further demonstrate that our proposed FNE can learn more accurate and discriminative feature representations. As mentioned in the main paper, our FNE can calculate the false negative probability for each negative sample, and assign low sampling weights to negative samples with high false negative probabilities to reduce the adverse impacts of false negatives during model optimization. This alleviates the contradiction between the optimization objectives of attracting positive samples and repelling false negatives, resulting in better performance in image-text matching. Therefore, compared to the model without the FNE strategy (w/o FNE), the attention between the target region and the target word will be more accurate and dense.

%%%%%%%%%%%%%%%%%==Table Block==%%%%%%%%%%%%%%%
\begin{table}[]
\centering
\caption{Investigation of the $\lambda$ in Equation~\ref{eq:eq_13}. Experiments are conducted on Flickr30K.} 
% \resizebox{\linewidth}{!}{
\setlength{\tabcolsep}{1.5mm}{
\begin{tabular}{c|cc|cc}
\hline
\multirow{2}{*}{$\lambda$} & \multicolumn{2}{c|}{Image-to-Text} & \multicolumn{2}{c}{Text-to-Image} \\ \cline{2-5} 
                       & R@1              & R@5             & R@1             & R@5             \\ \hline
0.5                    & 83.1             & 97.3            & 68.2            & 89.2            \\
0.3                    & 83.6             & 97.3            & 68.2            & 89.2            \\
0.1                    & 83.4             & 96.7            & 68.3            & 88.8            \\
0.01                   & 84.7             & 97.1            & 68.1            & 90.0            \\ \hline
\end{tabular}
}
\label{tab:lambda}
\end{table}
%%%%%%%%%%%%%%%%%==Table Block==%%%%%%%%%%%%%%%

\end{document}